\documentclass[10pt,a4paper,twoside]{article}
\usepackage[utf8]{inputenc}
\usepackage{lmodern}
\usepackage[T1]{fontenc}
\usepackage[english,spanish,activeacute]{babel}
\usepackage{url}
\usepackage[usenames]{xcolor}
\usepackage{algorithm2e}
\usepackage{./IEEEtrantools}
\usepackage{textcomp,fixmath} 
\usepackage{latexsym} 
\usepackage{makecell}
\usepackage{tablefootnote}
\usepackage{float}
\usepackage{amssymb}
\usepackage{amsmath} 
\usepackage{amsfonts}

\usepackage{pgfplots}
\usepackage{graphicx}
\usepackage{epstopdf} 
\usepackage{subcaption}
\usepackage[left=2.54cm, right=2.54cm, bottom=5.08cm,top=2.54cm]{geometry}

\usepackage{booktabs} 
\usepackage{multirow} 

\usepackage{fancyhdr}
\usepackage[backend=biber,sorting=none,url=false]{biblatex}
\usepackage[capitalise,noabbrev]{cleveref}
\Crefformat{figure}{#2Fig.~#1#3}
\Crefformat{equation}{#2Eq.~(#1)#3}
\Crefformat{table}{#2Table~#1#3}
\Crefformat{section}{#2Section~#1#3}

\RequirePackage[]{acronym}
\usepackage[xindy]{glossaries}
\newacronym{oms}{WHO}{World Health Organization}
\newacronym{fao}{FAO}{Food and Agriculture Organization of the United Nations}
\newacronym{cnn}{CNN}{Convolutional Neural Networks}
\newacronym{rcnn}{R-CNN}{Regression with Convolutional Neural Networks}
\newacronym{svm}{SVM}{Support Vector Machines}
\newacronym{pca}{PCA}{Principal Component Analysis}
\newacronym{sgdm}{SGDM}{Stochastic Gradient Descent with Momentum}
\newacronym{glcm}{GLCM}{Gray-Level Co-occurrence Matrix}
\newacronym{ann}{ANN}{Artificial Neural Networks}

\usepackage[range-phrase=-,
per-mode=symbol-or-fraction,
range-units=single,
list-units=single,
detect-all]{siunitx}
\DeclareSIUnit\decibeli{dBi}
\DeclareSIUnit\pixel{pixels}

\renewcommand{\tablename}{Table} 

\begin{document}
\selectlanguage{english}
\pagestyle{fancy}
\setcounter{page}{1}

\begin{center}
\large{\textbf{CNN-BASED SOLUTION FOR MANGO CLASSIFICATION IN AGRICULTURAL ENVIRONMENTS}}
\end{center}
\vspace{1em}

\begin{center}
	\textbf{Beatriz Díaz Peón\textsuperscript{1},
		Jorge Torres Gómez\textsuperscript{2},
		Ariel Fajardo Márquez\textsuperscript{3}}
\end{center}

\begin{center}
	\textsuperscript{1}Havana University of Technology ``José Antonio Echeverría'' (CUJAE), Havana, Cuba, \textsuperscript{2}TU Berlin, Berlin, Germany, \textsuperscript{3}Havana University of Technology ``José Antonio Echeverría'' (CUJAE), Havana, Cuba\\
	\textsuperscript{1}\url{eatrizdi@automatica.cujae.edu.cu},
	\textsuperscript{2}\url{jorge.torresgomez@tu-berlin.de}
	\textsuperscript{3}\url{arielfajardomarquez@gmail.com},
	
\end{center}

\selectlanguage{english}

\textbf{ABSTRACT}

This article exemplifies the design of a fruit detection and classification system using \gls{cnn}.
The goal is to develop a system that automatically assesses fruit quality for farm inventory management.
Specifically, a method for mango fruit classification was developed using image processing, ensuring both accuracy and efficiency.
Resnet-18 was selected as the preliminary architecture for classification, while a cascade detector was used for detection, balancing execution speed and computational resource consumption.
Detection and classification results were displayed through a graphical interface developed in MatLab App Designer, streamlining system interaction.
The integration of convolutional neural networks and cascade detectors proffers a reliable solution for fruit classification and detection, with potential applications in agricultural quality control.

\vspace{0.5em}\textbf{INDEX TERMS:} convolutional neural networks, MatLab, ResNet-18, cascade detector, regression with convolutional neural networks.

\selectlanguage{spanish}

SOLUCIÓN BASADA EN REDES NEURONALES CONVOLUCIONALES PARA LA CLASIFICACIÓN DE MANGOS EN ENTORNOS AGRÍCOLAS

\textbf{RESUMEN}

Este artículo ejemplifica el diseño de un sistema de detección y clasificación de frutas utilizando \gls{cnn}.
El objetivo es desarrollar un sistema que evalúe automáticamente la calidad de las frutas para la gestión del inventario agrícola.
En concreto, se desarrolló un método de clasificación para el mango utilizando el procesamiento de imágenes, garantizando tanto la precisión como la eficiencia.
Se seleccionó Resnet-18 como arquitectura preliminar para la clasificación, mientras que se utilizó un detector en cascada para la detección, equilibrando la velocidad de ejecución y el consumo de recursos computacionales.
Los resultados de la detección y la clasificación se mostraron a través de una interfaz gráfica desarrollada en MatLab App Designer, lo que agilizó la interacción con el sistema.
La integración de redes neuronales convolucionales y detectores en cascada ofrece una solución fiable para la clasificación y detección de frutas, con posibles aplicaciones en el control de calidad agrícola.

\vspace{0.5em}\textbf{PALABRAS CLAVES:} redes neuronales convolucionales, MatLab, Resnet-18, detector en cascada, regresión con redes neuronales convolucionales.

\vspace{1em}
\selectlanguage{english}

\section{INTRODUCTION}

Mango crops span most tropical and subtropical regions, with a global production exceeding $\SI{43e6}{\tonne}$, it is the third most produced and imported tropical fruit.
In the case of Cuba, mangoes hold significant relevance, representing $\SI{42}{\percent}$ of the cultivated fruit area and $\SI{30}{\percent}$ of overall fruit production~\cite{2cubamango}.
However, this fruit export is subject to strict international regulations, such as those outlined by the Codex Alimentarius,\footnote{Codex Alimentarius: \url{http://www.codexalimentarius.org/codexhome/en/}. Specifically, CODEX STAN 184-1993:  \url{www.codexalimentarius.org/download/standards/315/CXS\_184e.pdf}}, which establishes quality criteria including ripeness, classification, and contaminant control~\cite{alimentarius2005normas}.
Compliance with these standards poses a challenge for industries where fruit grading is performed manually, often resulting in subjective assessments and classification errors~\cite{2cubamango}.
To address this issue, it is essential to develop an automated system that ensures objective evaluation of mango quality.
As a solution, this work proposes the use of image processing techniques, specifically convolutional neural networks (\gls{cnn}).
This strategy enables precise fruit classification aligned with international quality standards.

Various image processing methods have proven effective for automated fruit classification.
For instance, approaches combining \gls{pca} and \gls{ann} have achieved approximately $\SI{79}{\percent}$ accuracy in detecting surface defects in apples~\cite{peterson2007identifying}.
Likewise, techniques that integrate k-means segmentation with feature extraction via \gls{glcm} have demonstrated outstanding results, achieving disease recognition rates in apples of nearly $\SI{99.60}{\percent}$~\cite{Sangeetha2018DetectionAC}.
However, these methods rely on manual feature extraction or selection algorithms that require expert knowledge, thereby limiting their generalization capacity under varying lighting conditions, camera angles, or image noise.

In contrast, \gls{cnn} architectures offer significant advantages for classification and detection tasks, as they automatically extract hierarchical image features, ranging from simple patterns to complex structures~\cite{barba2021uso}.
Their basic components include convolutional, pooling, and fully connected layers, with activation functions such as Softmax,\footnote{Calculates the probability of each class relative to all possible classes.} and ReLU,\footnote{Enables the model to approximate nonlinear functions by linking pixels with the network's semantic content.} which optimize learning and prediction processes~\cite{barba2021uso}.

The use of pretrained networks such as ResNet-50, ResNet-18, AlexNet and GoogLeNet, available through the MatLab Deep Learning Toolbox, facilitates adaptation to new tasks via transfer learning~\cite{Deep_Toolbox}.
For problems that require both classification and localization, architectures such as \gls{rcnn} and its optimized variants (Fast \gls{rcnn}, Faster \gls{rcnn}, and Mask \gls{rcnn}) are employed~\cite{matlab_rcnn}.
Cascaded detectors are also used to discard irrelevant regions and accelerate processing, which is particularly useful for real-time applications~\cite{matlab_cascade}.
These detectors focus only on regions with a high probability of containing relevant information, providing a viable solution for real-time environments.

These architectures have demonstrated high precision levels in complex environments.
Deep models such as VGG and ResNet-50 have shown accuracies between $\SI{92}{\percent}$ and $\SI{93}{\percent}$ in detecting orange diseases~\cite{agrawal2020orange} and up to $\SI{99.6}{\percent}$ in strawberry disease classification~\cite{xiao2020detection}.
Similarly, the Faster R-\gls{cnn} model, which combines region proposals with the robustness of \gls{cnn}, has proven effective in detecting blueberry defects for export classification~\cite{horna2021sistema}.
These capabilities justify the selection of \gls{cnn}-based models as the foundation of the proposed system.

Following this trend, the main objective of this research is to develop a system based on \gls{cnn} models for mango detection and classification.
To this end, suitable architectures were selected, preprocessing algorithms were applied to the datasets and hyperparameters\footnote{Configurations that are not learned directly during training but must be set beforehand or during network construction.} were tuned through experimental trials.
Additionally, a graphical interface was developed using MatLab App Designer to integrate the trained models, enabling practical use in real environments.

The obtained results validate the effectiveness of the proposed methodology.
The Resnet-18 architecture achieved an accuracy of $\SI{89.51}{\percent}$ for ripeness classification and $\SI{90.65}{\percent}$ for disease classification.
These figures indicate strong performance both in assessing mango maturity and in identifying potential diseases.
Moreover, the cascaded detector reached an accuracy of $\SI{90}{\percent}$ in mango detection within images, confirming its applicability in industrial contexts.
It was also observed that excessively deep networks may lead to overfitting,\footnote{
\url{https://enclavedeciencia.rae.es/sobreajuste}} which impairs model generalization.
Thus, smaller and more task-specific architectures were favored.

Finally, different detection models were evaluated.
Although \gls{rcnn}-based architectures offered good precision, their high execution time makes them less suitable for real-time applications and systems with limited computing resources.
In contrast, the cascaded detector showed a better balance between speed and accuracy, making it more appropriate for resource-constrained environments.

The rest of the article continues as follows: In \Cref{sec_materiales}, the methodology used to create the models is presented.
In \Cref{sec_resultados}: , the performance of the proposed system is discussed.
Finally, the last section concludes the article, highlighting the main features of this proposal.

\section{MODEL DEVELOPMENT FOR MANGO CLASSIFICATION AND DETECTION}
\label{sec_materiales}

This section outlines the methodology used to develop the mango detection and classification system.
It details the selection and preprocessing of the datasets, as well as the architectures employed for each task.
For detection, models based on \gls{rcnn} and cascade detectors were used, while classification was addressed through \gls{cnn} architectures adapted for identifying ripeness stages and diseases.
Modifications made to the architectures and the training process are also explained.

\subsection*{Dataset search and selection for training}

Selecting an appropriate dataset is essential for effective neural network training, as it directly impacts model accuracy and generalization.
In this research, datasets were analyzed according to three specific tasks: 1) classification of mango ripeness, 2) disease classification, and 3) mango detection in images.
Since each task requires different label types and structural formats, a single dataset could not meet all requirements.
Therefore, a comprehensive search was conducted across repositories such as Roboflow,\footnote{\url{https://universe.roboflow.com/}} GitHub,\footnote{\url{https://github.com/}} and Mendeley Data,\footnote{\url{https://data.mendeley.com/}} prioritizing datasets with sufficient volume, visual diversity and well-defined labels.

After selecting a general set of datasets for the various tasks, three relevant options were evaluated and compared for mango ripeness classification.
The considered datasets were: ``Mango detection system yolov8.v2-yolov8-new-dataset'',\footnote{\url{https://universe.roboflow.com/mango-yolov8/mango-detection-system-yolov8-mcocq}} ``Mangos.v6i'',\footnote{\url{https://universe.roboflow.com/mango-detection/mangos-kyo1z}} and ``Mango Variety and Grading Dataset''.\footnote{\url{https://data.mendeley.com/datasets/5mc3s86982/1}}.
\Cref{tb1} summarizes their main characteristics.

\renewcommand{\tablename}{Table}
\begin{table}[htb] 
	\caption{
		\label{tb1} Comparison of datasets for mango ripeness classification.} 
	\begin{center} 
		\begin{tabular}{|c|c|c|c|c|} 
			\hline 
			Dataset & \makecell{Number of \\categories} & \makecell{Number of \\images} & \makecell{Noise \\conditions} & \makecell{Lighting and \\background variations} \\ 
			\hline 
			\makecell{Mango detection system\\yolov8.v2-yolov8-new-dataset} & \num{3} & \num{17009} & Included & Included \\ 
			\hline 
			Mangos.v6i & \num{3} & \num{8953} & Not included & Limited\\ 
			\hline 
			\makecell{Mango Variety and \\Grading Dataset} & \num{11} & \num{2200} & Not included & Not included \\ 
			\hline 
		\end{tabular} 
	\end{center} 
\end{table}

\begin{figure}[htb]
	\centering
	\subfloat[Bad mango]
	{\includegraphics[width=0.25\columnwidth]{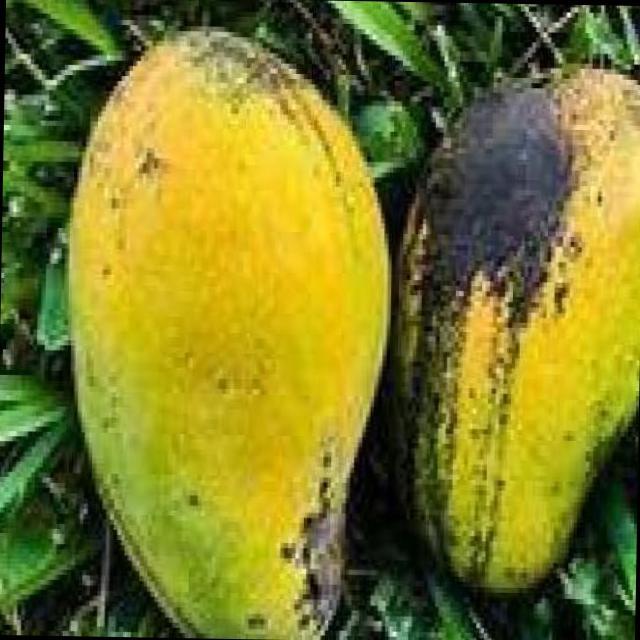}}
	\hfill
	\subfloat[Raw mango]
	{\includegraphics[width=0.25\columnwidth]{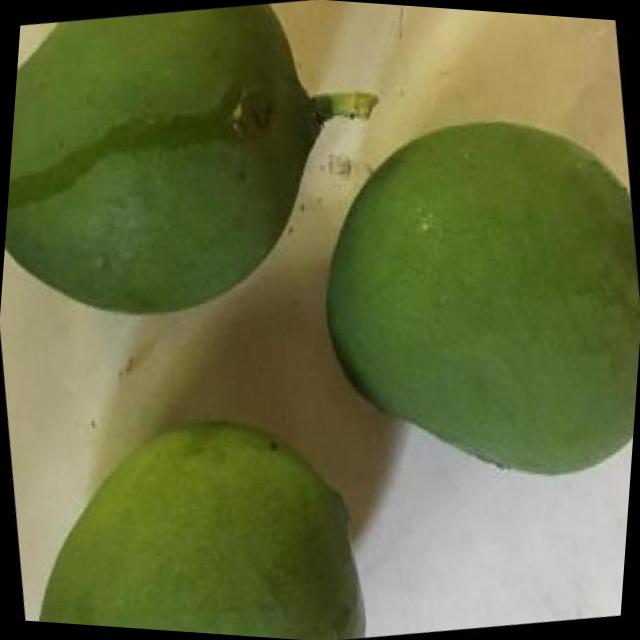}}
	\hfill
	\subfloat[Ripe mango]
	{\includegraphics[width=0.25\columnwidth]{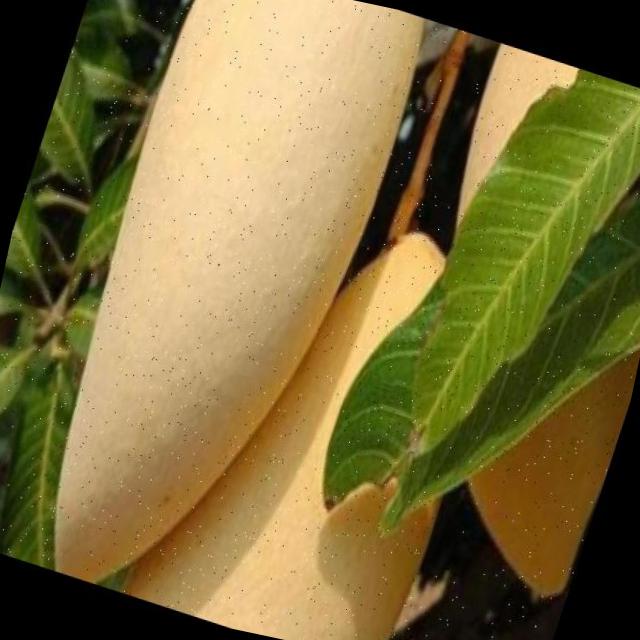}}
	\caption{Digital images of the categories: bad mango, raw mango and ripe mango.}
	\label{fig1_2_3}
\end{figure}

Based on this analysis, the ``Mango detection system yolov8.v2-yolov8-new-dataset'' was selected.
Its high resolution and diversity of lighting and background conditions contribute to stronger model performance in real-world environments.
Moreover, it is considerably larger than the other datasets evaluated, allowing better class representation and minimizing the risk of overfitting. 
It contains \num{17009} images at \mbox{$\SIrange[range-phrase=\times]{640}{640}{\pixel}$} resolution, divided into three categories: bad mango (\num{6856} images), raw mango (\num{4772} images) and ripe mango (\num{4672} images), as shown in \Cref{fig1_2_3}. Notably, this dataset includes three variations per image.
Approximately $\SI{50}{\percent}$ of the images are flipped either vertically or horizontally, randomly rotated between $\SI{-15}{\degree}$ and $\SI{+15}{\degree}$ and subjected to random Gaussian blur in the range of $\SIrange{0}{2.3}{\pixel}$.
These transformations, integrated into the dataset, improve model robustness against visual variations.

For mango disease classification, three datasets were identified and evaluated independently of the one used for ripeness classification, as they address a distinct task within the study.
The datasets considered were: ``project > 2023-12-04 1:29pm'',\footnote{\url{https://universe.roboflow.com/project-khw09/project-airhk}} ``Mango.v6i.multiclass'',\footnote{\url{https://universe.roboflow.com/cocoa-f6c3y/mango-qhzr4}} and ``mangos.v7i.multiclass''.\footnote{\url{https://universe.roboflow.com/yolo-pzqub/mangos-zjjq7}}
\Cref{tb4} summarizes them based on the number of classes, image volume, and disease representativity.

\begin{table}[htb] 
	\caption{
		\label{tb4} Comparison of datasets for mango disease classification.} 
		\begin{center} 
			\begin{tabular}{|c|c|c|c|c|} 
				\hline Dataset & \makecell{Number of \\categories} & \makecell{Number of \\images} & \makecell{Noise \\conditions} & \makecell{Lighting and \\background variations}\\ 
				\hline project > 2023-12-04 1:29pm & \num{5} & \num{4034} & Yes & Yes\\ 
				\hline Mango.v6i.multiclass & \num{5} & \num{9825} & No & Yes\\ 
				\hline mangos.v7i.multiclass & \num{3} & \num{485} & No & No \\ 
				\hline 
			\end{tabular} 
		\end{center} 
	\end{table}

\begin{figure}[htb]
	\centering
	\subfloat[Alternaria]
	{\includegraphics[width=0.25\columnwidth]{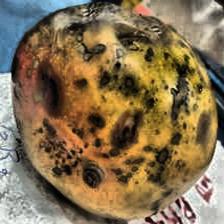}}
	\hfill
	\subfloat[Anthracnose]
	{\includegraphics[width=0.25\columnwidth]{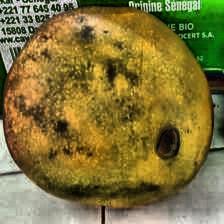}}
	\hfill
	\subfloat[Black mold rot]
	{\includegraphics[width=0.25\columnwidth]{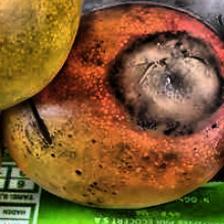}}
	\\
	\subfloat[Healthy]
	{\includegraphics[width=0.25\columnwidth]{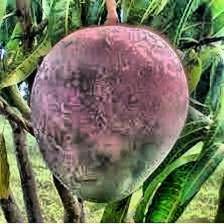}}
	\hfill
	\subfloat[Stem end rot]
	{\includegraphics[width=0.25\columnwidth]{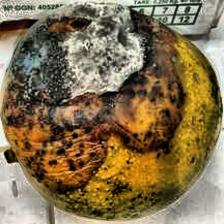}}
	\caption{Digital images of the categories: alternaria, anthracnose, black mold rot, healthy and stem end rot.}
	\label{fig10_11_12_13_14}
\end{figure}

Among the analyzed datasets, ``project > 2023-12-04 1:29pm'' was selected due to its wide range of image variants.
It includes \num{4034} images with $\SIrange[range-phrase=\times]{640}{640}{\pixel}$ resolution, divided into five categories: alternaria (\num{795} images), anthracnose (\num{614} images), black mold rot (\num{877} images), healthy (\num{980} images) and stem end rot (\num{766} images), as illustrated in \Cref{fig10_11_12_13_14}.
This dataset also contains three image variants per sample.
The transformations applied include horizontal flipping ($\SI{50}{\percent}$), random rotation of $\SI{90}{\degree}$ (none, clockwise, or counterclockwise), random cropping of up to $\SIrange{0}{20}{\percent}$ and Gaussian blur within $\SIrange{0}{2.5}{\pixel}$.
These embedded variants prepare the model for real-world conditions, where images may exhibit noise, slight blur, or rotation.

\begin{figure}[htb]
	\centering
	\subfloat[Mango]
	{\includegraphics[width=0.25\columnwidth]{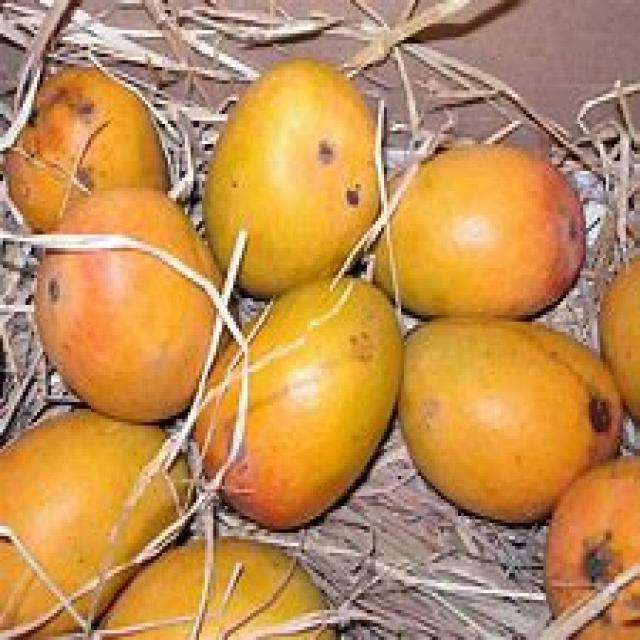}}
	\hfill
	\subfloat[Banana]
	{\includegraphics[width=0.25\columnwidth]{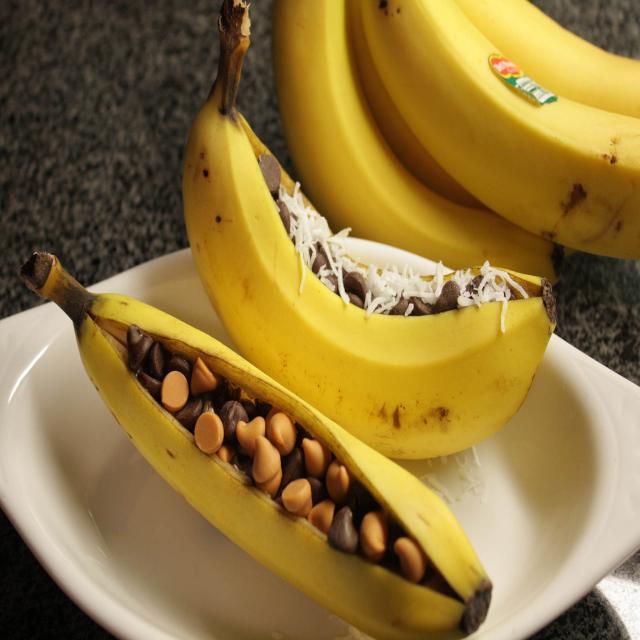}}
	\hfill
	\subfloat[Avocado]
	{\includegraphics[width=0.25\columnwidth]{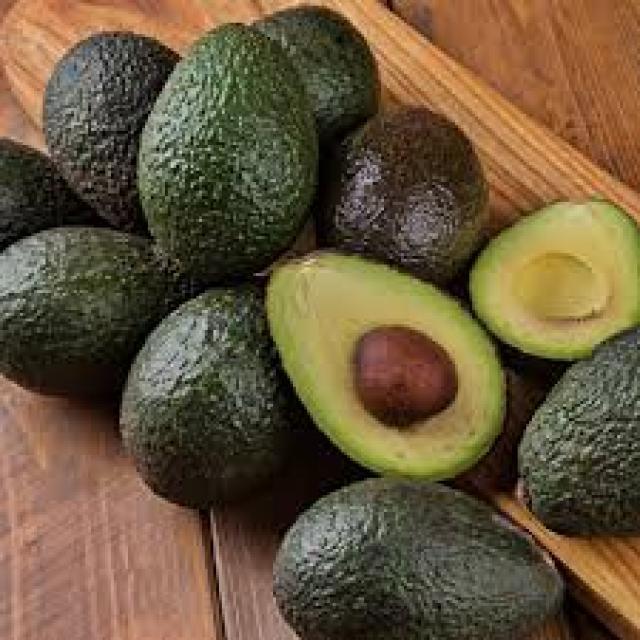}}
	\caption{Digital images from different categories: mango, banana and avocado.}
	\label{fig7_8_9}
\end{figure}

Finally, for mango detection in images, the dataset ``572 FRUITS VEGETABLES.v1i.tensorflow''\footnote{\url{https://universe.roboflow.com/ml-datatrain/572\_fruits\_vegetables}} was selected.
It consists of \num{11489} images at $\SIrange[range-phrase=\times]{640}{640}{\pixel}$ resolution. Unlike the previous datasets, it is not folder-organized but includes a \texttt{CSV} file specifying each image’s filename, width, height, class, and bounding box coordinates.
The existing categories include mango, strawberry, banana, guava, tomato, onion, egg, avocado, cherry, peach, apple, pineapple, kiwi, among others.
Some examples are shown in \Cref{fig7_8_9}.
As an alternative, the dataset ``FruitMate.v2i.coco''\footnote{\url{https://universe.roboflow.com/tharushi-qpqyp/fruitmate}} was considered.
It contains \num{5977} images and \num{10} categories.
However, its smaller size and limited variability in background and lighting conditions could hinder model training for classification in real environments.
Thus, it was discarded in favor of the first dataset.

\subsection*{Data preprocessing for neural network training}

Data preprocessing is a crucial step to ensure image compatibility with the selected models.
This process involves several fundamental operations:

\begin{enumerate} 
	\item Dataset splitting: Images are divided into training and validation sets (MatLab recommends a $\SIrange[range-phrase=-]{70}{30}{\percent}$ split)\footnote{This division can be adjusted based on the problem.
	Validation should not significantly exceed the training set, as it may slow down the model’s learning process.}~\cite{matlab_cnn}.
	This separation can be performed in two ways: 
	\begin{itemize} 
		\item Random/Proportional Sampling: Random selection of images or selection based on predefined proportions from the original dataset. 
		\item Predefined Structure: Use of organized folders such as \texttt{train}, \texttt{valid} and \texttt{test}, if the dataset already follows that structure.
	\end{itemize}
	
	\item Image resizing: Images are adjusted to the standard input sizes required by the pretrained models used ($\SIrange[range-phrase=\times]{224}{224} {\pixel}$ and $\SIrange[range-phrase=\times]{227}{227} {\pixel}$), such as ResNet-50 , GoogLeNet and AlexNet~\cite{matlab_cnn_pretrained}.
	This step ensures compatibility and computational efficiency.
	The original dataset structure is preserved during this process to facilitate handling.
	
	\item Bounding box adjustment (for object detection): In detection tasks, it is necessary to modify the~$xy$ coordinates of the bounding boxes to maintain their proportions relative to the new image size.
	This ensures that the regions of interest continue to correspond correctly to the detected objects. 
\end{enumerate}

Finally, the MatLab Image Labeler tool was used to verify the correct modification of both images and bounding boxes.
This tool allows manual inspection and correction of alignment errors or missing labels, although the process can be time-consuming.

\subsection*{Architectures for mango classification and detection}

\begin{figure}[htb]
	\centering
	{\includegraphics[trim=50 35 45 35, clip,width=0.95\columnwidth]{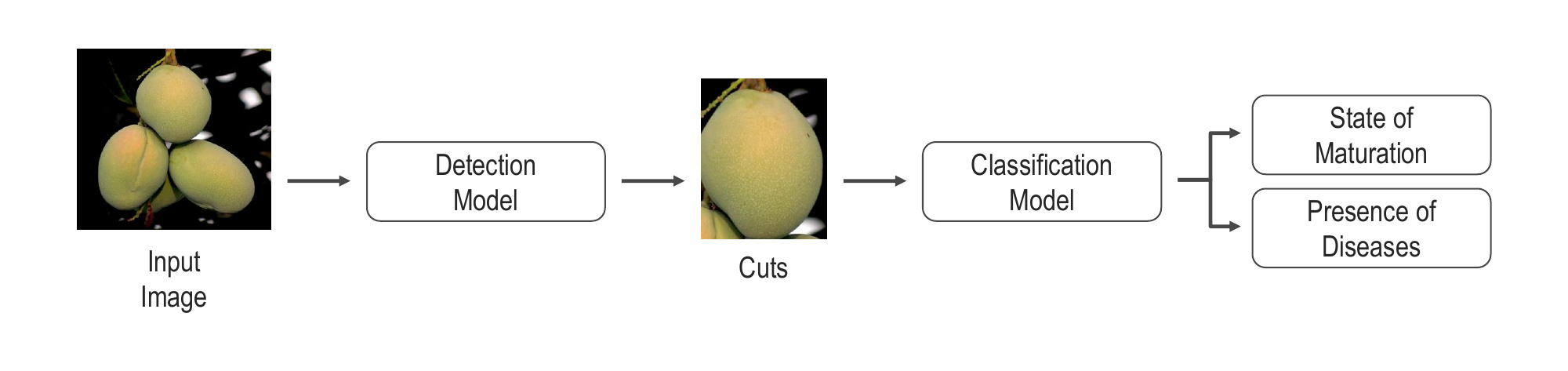}}
	\caption{Workflow of the mango detection and classification system.}
	\label{fig18}
\end{figure}

The proposed solution consists of two sequential stages, as illustrated in \Cref{fig18}.
First, all mangoes present in the whole image are detected, generating individual crops for each detection.
Then, each crop is analyzed separately during the classification stage, where both ripeness and disease presence are determined.
This workflow ensures that the system processes only regions validated as mangoes, optimizing computational resources and enabling more precise object-level analysis.

\begin{figure}[htb]
	\centering
	{\includegraphics[trim=15 30 30 30, clip,width=0.95\columnwidth]{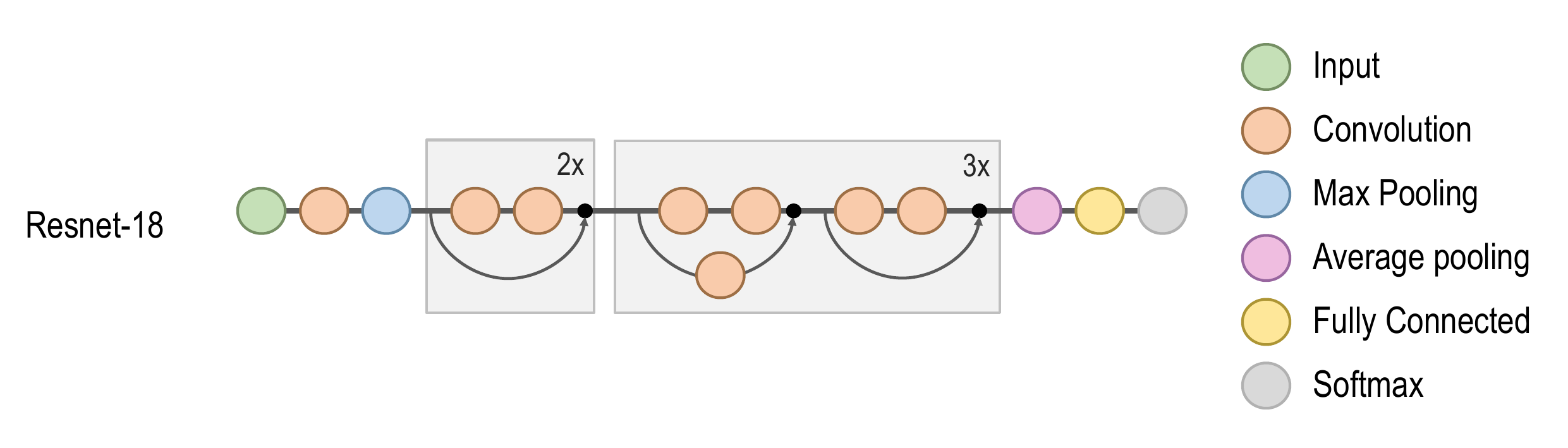}}
	\caption{Representation of the Resnet-18 architecture.}
	\label{fig19}
\end{figure}

To classify ripeness and diseases, the pretrained Resnet-18 architecture was used.
It comprises 18 layers, including convolutions, batch normalization, and ReLU activation functions.
Its residual block structure,\footnote{These blocks directly connect the output of specific layers with subsequent layers.} facilitates gradient flow during training, enhancing model stability and accuracy~\cite{resnet18}. 
\Cref{fig19} shows a schematic representation of this architecture.

As Resnet-18 was initially designed to classify up to \num{1000} categories, its final layers were modified to match the number of classes required for this problem~\cite{matlab_cnn_pretrained}.
These layers combine extracted features from the convolutional blocks to generate class probabilities, compute the loss function and produce predicted labels.
In ResNet-18, the last trainable layer is fully connected and it was replaced with a new one containing the appropriate number of output nodes (\num{3} for ripeness, \num{5} for disease detection).
Additionally, the classification layer was replaced so that the model could learn the dataset labels directly~\cite{matlab_cnn}.
This adaptation ensures that Resnet-18 is optimized for the specific classification tasks presented.

\begin{figure}[htb]
	\centering
	{\includegraphics[trim=22 10 30 10, clip,width=0.9\columnwidth]{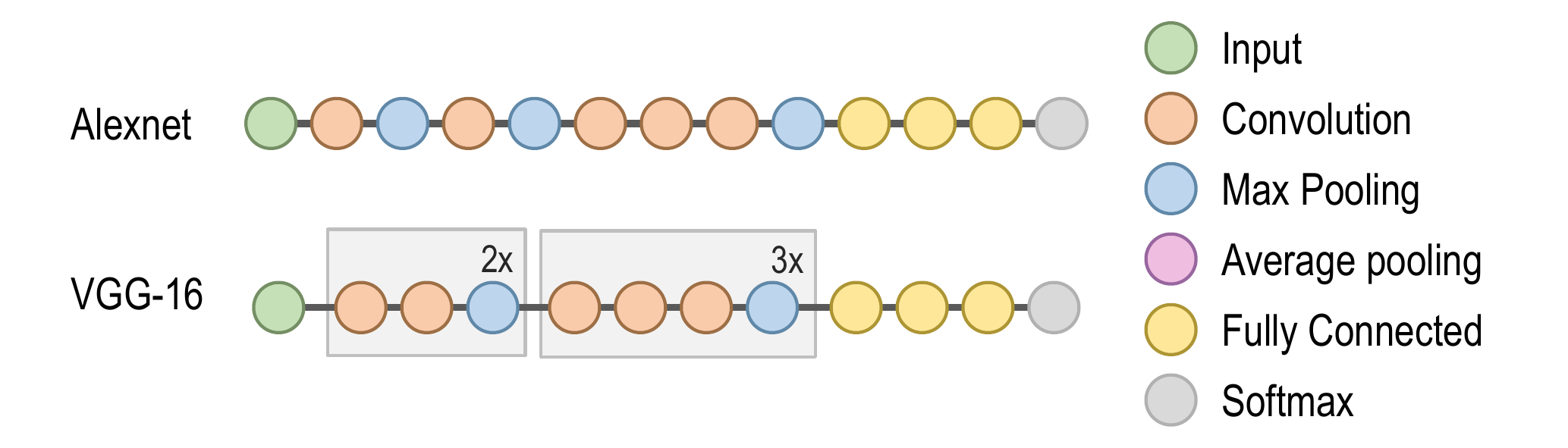}}
	\caption{Representation of the AlexNet and VGG-16 architectures.}
	\label{fig20}
\end{figure}

For detection, two classic CNN architectures—AlexNet and VGG-16—were evaluated and modified, as illustrated in \Cref{fig20}.
AlexNet consists of \num{5} convolutional layers, \num{3} pooling layers, \num{3} fully connected layers and a Softmax layer~\cite{alexnet}.
VGG-16, on the other hand, has \num{16} layers in total (\num{13} convolutional and \num{3} fully connected), uses \numrange[range-phrase=$\times$]{3}{3} convolutions with a large number of filters and contains approximately \num{138e6} parameters~\cite{vgg16}.

Both networks were adapted to the \gls{rcnn} framework by removing their original final layers (fully connected, Softmax and classification), which were replaced with new layers configured to classify proposed regions, including an additional background class.
This process is similar to the one used to adapt \gls{cnn} to new datasets. 
The strategy retains the pretrained convolutional layers for feature extraction, while the added layers specialize in detecting relevant mango-related objects~\cite{matlab_rcnn_detection}.

In addition to \gls{rcnn}-based models, a cascade object detector was employed to provide a lightweight alternative for real-time environments.
The cascade detector is particularly effective for recognizing object categories with relatively stable aspect ratios (faces, traffic signs, and vehicles).
It operates using a hierarchical architecture that classifies image windows through multiple stages.
Each stage quickly eliminates regions unlikely to contain the target object, allowing the system to focus computational resources on the most promising areas.
At every step, object-specific features are evaluated to determine whether the region advances to the next classifier or is discarded~\cite{matlab_cascade}.

\subsection*{Training for detection and classification}

Two approaches were employed for object detection in images: the \gls{rcnn} method using pretrained networks (VGG-16 and AlexNet) and cascade detector.
The \gls{cnn} architectures were adjusted to create an \texttt{rcnnObjectDetector} object through an initial training phase.
Final training was conducted using the \gls{sgdm} solver, with a mini-batch size of \num{32}, over \num{10} epochs and an initial learning rate of \num{0.000001}.

In the case of the cascade detector, different parameter configurations were evaluated, as shown in \Cref{tb9}.
Initial tests produced multiple false detections; however, by increasing the value of \texttt{NumCascadeStage},\footnote{Number of cascade stages: Specifies how many stages the cascade detector applies.} and reducing the \texttt{FalseAlarmRate},\footnote{False alarm rate: Indicates the percentage or frequency of incorrect detections.} detection accuracy significantly improved in Tests 5 and 6.
Ultimately, the configuration in Test 6 (\texttt{FalseAlarmRate} of \num{0.05}, \texttt{NumCascadeStage} of \num{10}) was selected for providing the best balance between precision and computational efficiency.

\begin{table}[htb]
	\caption{\label{tb9} Training parameters of the cascade detector.}
	\begin{center}
		\begin{tabular}{|c|c|c|c|c|c|c|}
			\hline
			Hyperparameter & Test 1 & Test 2 & Test 3 & Test 4 & Test 5 & Test 6\\
			\hline
			FalseAlarmRate & \num{0.01} & \num{0.1} & \num{0.2} & \num{0.05} & \num{0.15} & \num{0.05} \\
			\hline
			NumCascadeStage & \num{10} & \num{5} & \num{15} & \num{5} & \num{20} & \num{10} \\
			\hline
			ObjectTrainingSize & auto & [24 24] & auto & auto & auto & auto \\
			\hline
		\end{tabular}
	\end{center}
\end{table}

For ripeness classification (raw, ripe, and rotten), the Resnet-18 architecture described earlier was trained.
A total of \num{14884} images were used for training, \num{1416} for validation and \num{709} for testing.
To balance the training and validation data, \num{750} images per class were randomly selected for training and~\num{100} images per class for validation.
This strategy contributed to a robust model, enabling accurate performance evaluation.

Various hyperparameter configurations were tested, as summarized in \Cref{tb7}.
The final model was trained using the \gls{sgdm} solver, with an initial learning rate of \num{0.001}, a mini-batch size of \num{32} and \num{10} epochs, applying a progressive learning rate drop of \num{0.01} per epoch.

\begin{table}[htb]
	\caption{\label{tb7} Hyperparameter configuration.}
	\begin{center}
		\begin{tabular}{|c|c|c|c|c|c|}
			\hline
			Hyperparameter & Test 1 & Test 2 & Test 3 & Test 4 & Test 5\\
			\hline
			InitialLearnRate & \num{0.001} & \num{0.01} & \num{0.01} & \num{0.001} & \num{0.001}\\
			\hline
			LearnRateSchedule & not applied & piecewise & piecewise & not applied & piecewise\\
			\hline
			LearnRateDropPeriod & not applied & \num{3} & \num{1} & not applied & \num{4}\\
			\hline
			LearnRateDropFactor & not applied & \num{0.1} & \num{0.01} & not applied & \num{0.1}\\
			\hline
			L2Regularization & not applied & not applied & not applied & \num{0.01} & \num{50}\\
			\hline
		\end{tabular}
	\end{center}
\end{table}

Once the ripeness was classified, the next step was to identify possible diseases.
The ``spoiled'' category includes deteriorated fruits regardless of ripeness, as damage may result from either senescence or disease.
Therefore, further analysis is necessary to identify the specific cause and assess the risk of total crop loss.
Following experimentation, two approaches were considered:

\begin{enumerate}
	\item Combined multi-category model: A single network for ripeness and disease (higher complexity, lower precision).
	\item Dedicated model for disease classification: A separate network for disease detection (more interpretable and accurate).
\end{enumerate}

Since the first approach resulted in lower precision and higher ambiguity (e.g., mistaking ripening spots for pathological symptoms), the second alternative was chosen.
Resnet-18 was again employed, adapted to five disease categories.
A total of \num{3537} images were used for training, \num{331} for validation and \num{166} for testing.
A total of \num{300} images per class were randomly selected for training and \num{50} per class for validation.
Training was performed using hyperparameters similar to those used in the previous stage: the \gls{sgdm} solver, initial learning rate of \num{0.01}, mini-batch size of \num{32} and \num{10} epochs, with a learning rate reduction of \num{0.01} to enhance convergence.

\section{RESULT ANALYSIS}
\label{sec_resultados}

This section presents the results obtained in the detection and classification of mangoes according to their ripeness and disease status.
For object detection, the dataset ``572 FRUITS VEGETABLES.v1i.tensorflow'' was used, evaluating both \gls{rcnn} models (based on VGG-16 and AlexNet) and cascade detectors.
Classification tasks were addressed using Resnet-18 models, trained with the datasets ``Mango detection system yolov8.v2-yolov8-new-dataset'' (for ripeness) and ``project > 2023-12-04 1:29pm'' (for diseases).
Additionally, precision, stability, and computational efficiency metrics were analyzed to determine the most suitable configurations.

\subsection*{Training and validation of the mango detection model}

Two \gls{rcnn}-based models—VGG-16 and AlexNet—were evaluated.  
Both training processes were manually stopped to prevent overfitting due to excessive iterations.  
\Cref{tb10} summarizes the main results obtained for the training phase.
It can be observed that both models exhibit quite similar accuracy, while the VGG-16 model excels with the lowest training loss.

\begin{figure}[htb]
	\centering
	\subfloat[AlexNet]
	{\includegraphics[trim=0 0 0 12, clip,width=0.3\columnwidth]{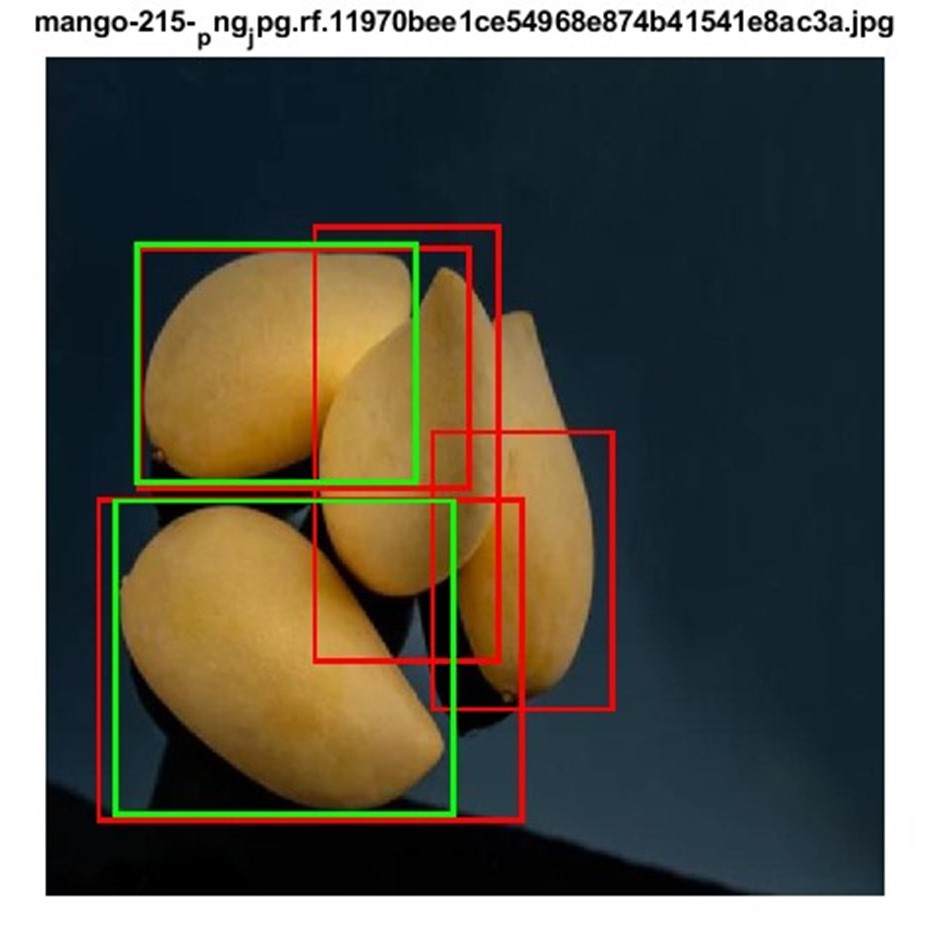}}
	\qquad
	\subfloat[VGG-16]
	{\includegraphics[trim=0 0 0 12, clip,width=0.3\columnwidth]{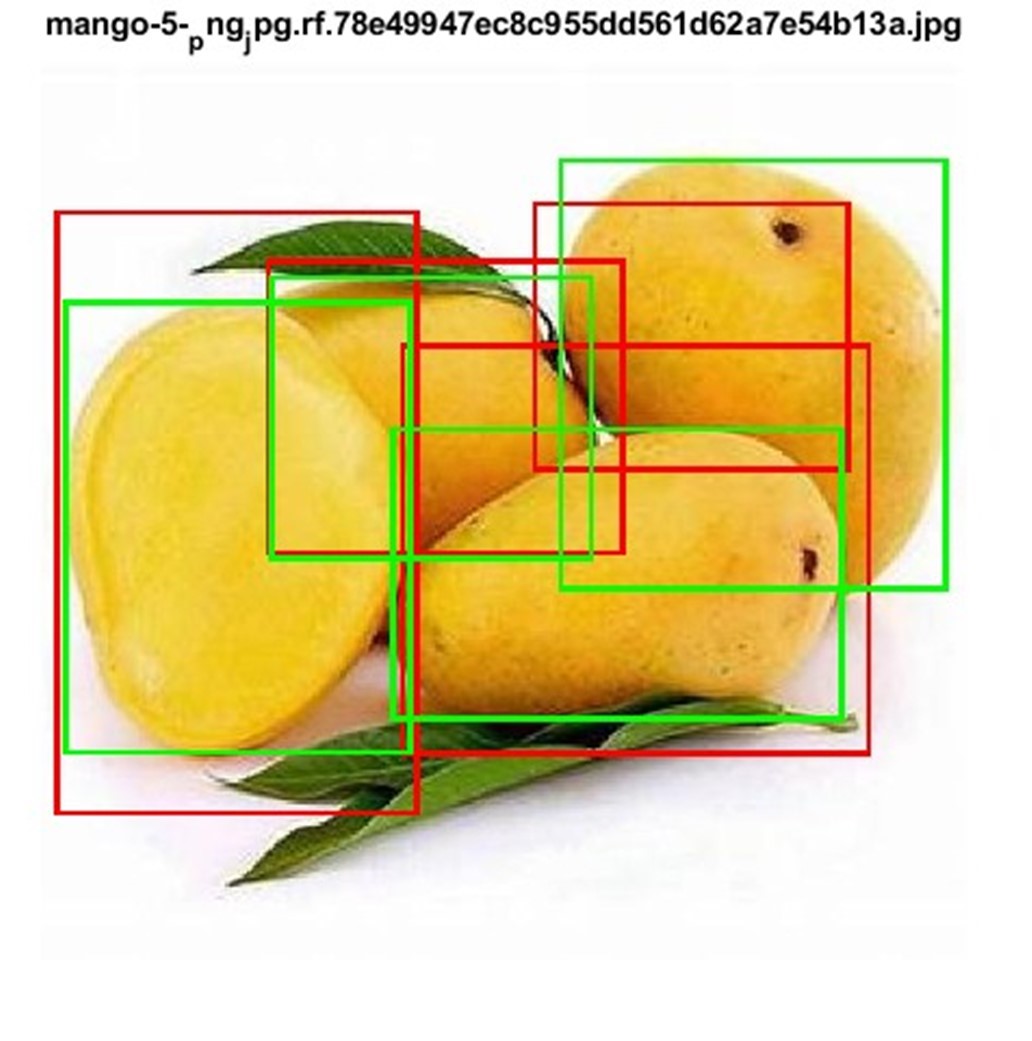}}
	\caption{Mango detection using \gls{rcnn}-based models: AlexNet and VGG-16}
	\label{fig31_32}
\end{figure}

Using these two models, detection results for each model are illustrated in \Cref{fig31_32}.
Green boxes indicate ground truth, while red boxes represent predictions generated by the models.
By comparing both results, a better correspondence is observed for the VGG-16 model compared to AlexNet.

\begin{table}[htb]
	\caption{\label{tb10} Training results of \gls{rcnn}.}
	\begin{center}
		\begin{tabular}{|c|c|c|c|c|}
			\hline
			Models & Training Accuracy (\%)  & Training Loss & Training Time & Iteration\\
			\hline
			VGG-16 & $\SI{79.00}{\percent}$ & \num{0.75} & $\SI{417}{\minute}$ $\SI{7}{\second}$ & \numrange[range-phrase=~from~]{597}{7490}\\
			\hline
			AlexNet & $\SI{80.00}{\percent}$ & \num{0.5} & $\SI{100}{\minute}$ $\SI{54}{\second}$ & \numrange[range-phrase=~from~]{1695}{7490}\\
			\hline
		\end{tabular}
	\end{center}
\end{table}

AlexNet demonstrated good generalization capability, detecting even unlabeled objects, though with slight misalignments in some bounding boxes.  
VGG-16, due to its greater depth, achieved better alignment of detections, albeit with higher computational cost.  
Both models produced acceptable results, balancing precision, speed and complexity.

\begin{figure}[htb]
	\centering
	\includegraphics[trim=0 12 0 15, clip,width=0.6\columnwidth]{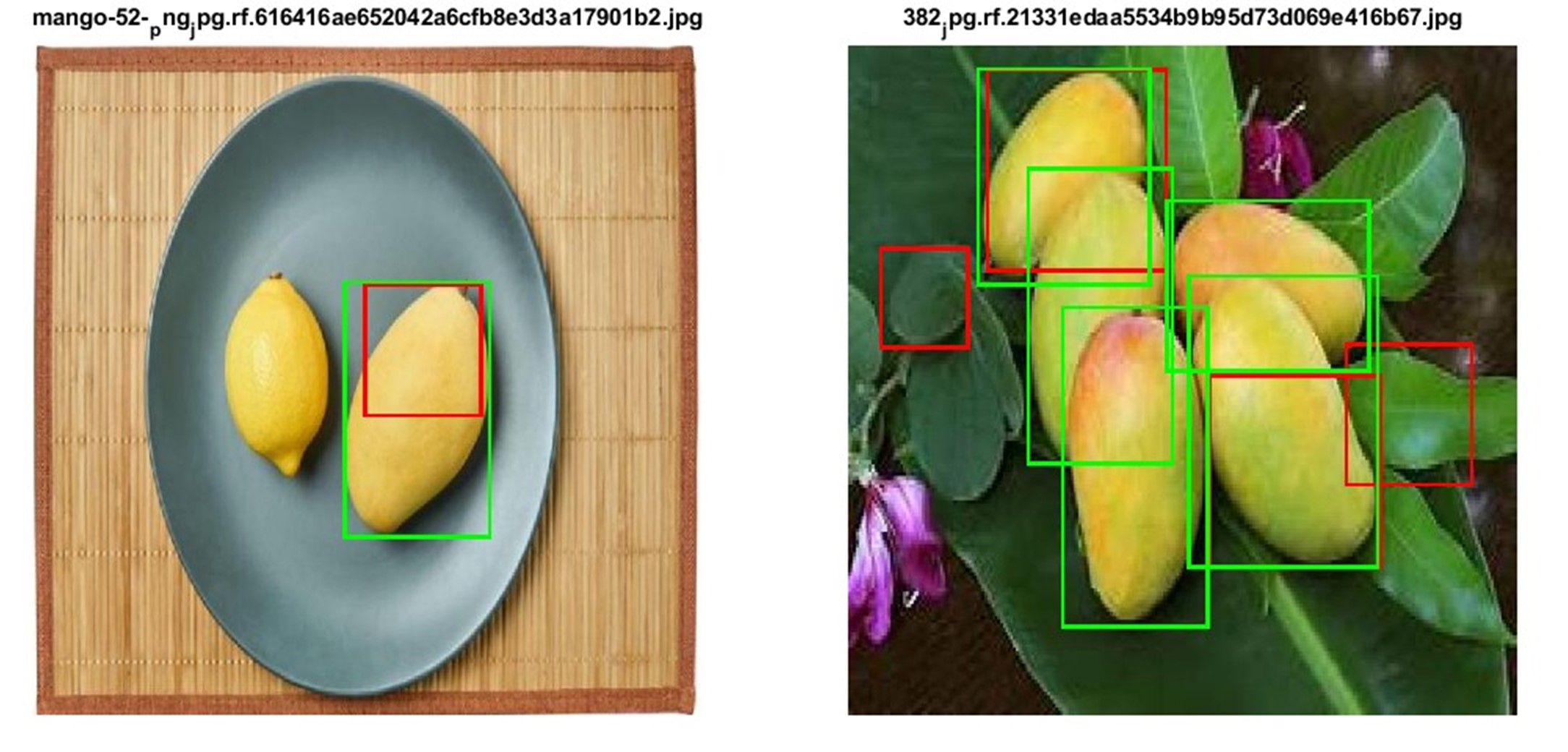}
	\caption{Mango detection in initial tests with the cascade detector}
	\label{fig33}
\end{figure}

\begin{figure}[htb]
	\centering
	\includegraphics[trim=0 12 0 15, clip,width=0.6\columnwidth]{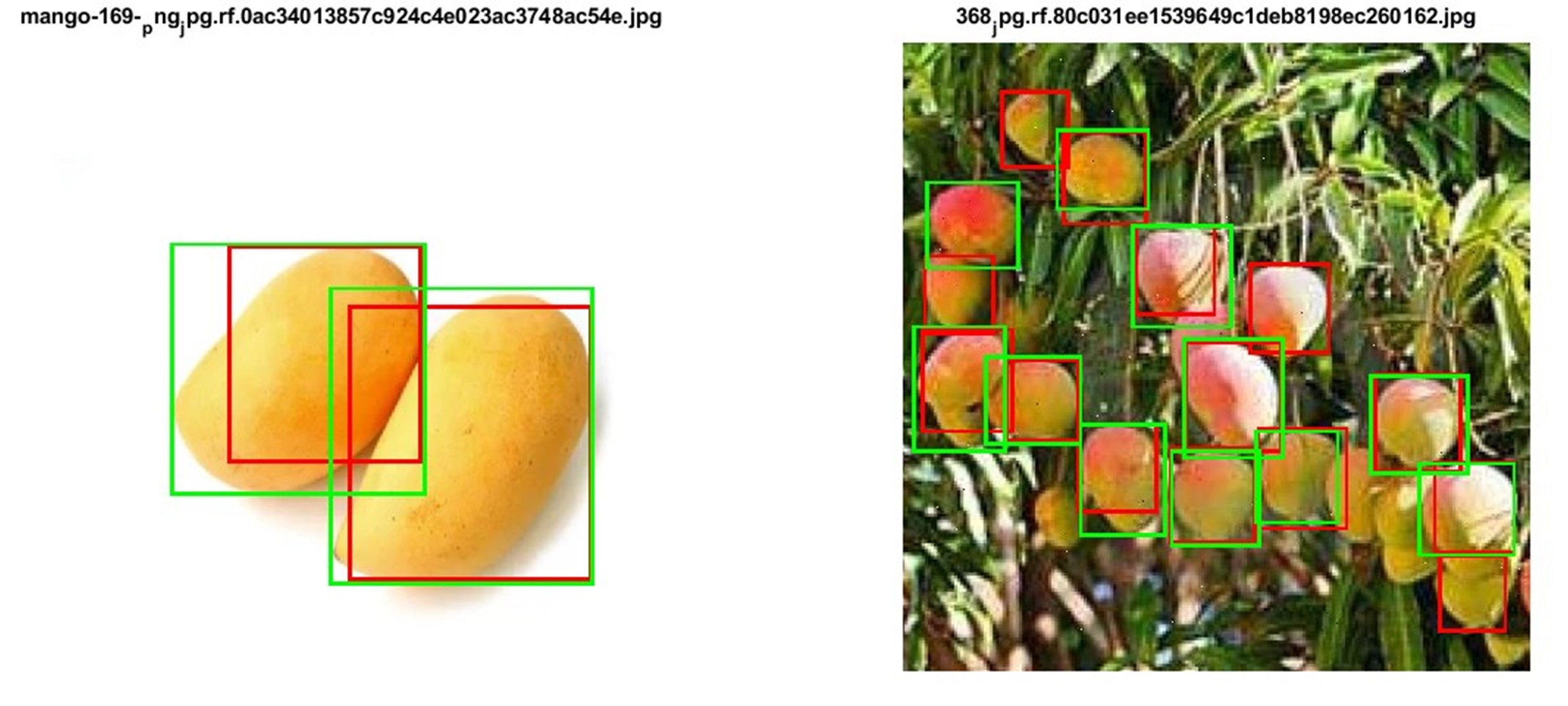}
	\caption{Mango detection in images using the cascade detector}
	\label{fig34}
\end{figure}

As a lighter alternative, a cascade object detector was evaluated.  
\Cref{fig33} shows that initial testing produced unsatisfactory results, with multiple incorrect detections and irrelevant regions, indicating low discrimination capability.
However, after adjusting the parameters, by increasing the number of stages and lowering the false alarm rate, accuracy improved significantly, as illustrated in \Cref{fig34}.  
Predicted boxes were better aligned with objects of interest, and false detections on leaves and other structures were notably reduced.

The cascade detector was ultimately selected for its simplicity, speed, and strong performance even with limited training data.  
Its ability to operate in real time and under variable lighting and scale conditions makes it ideal for practical applications.  
Although \gls{rcnn} models offer higher precision in complex environments, their high computational cost makes them less suitable for resource-constrained systems.

\subsection*{Training and validation of the ripeness classification model}

\begin{figure}[htb] 
	\centering 
	\subfloat[Ripeness classification] {\includegraphics[width=0.4\columnwidth]{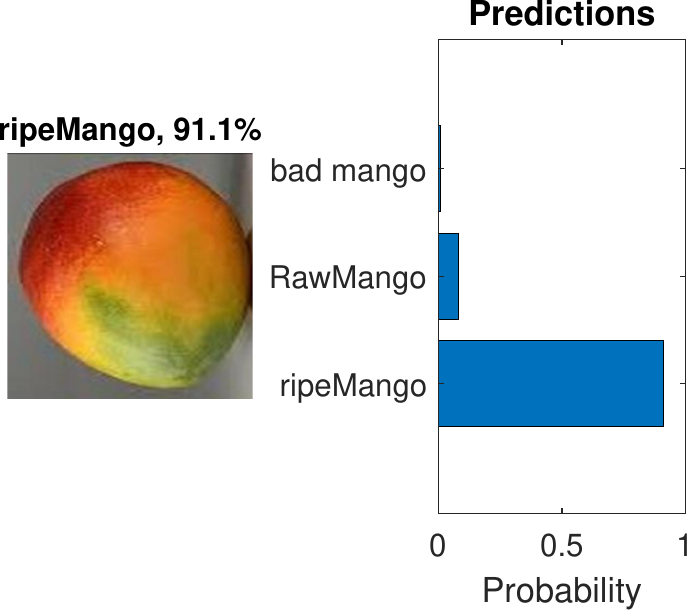}} 
	\hfill
	\subfloat[Model confusion matrix]
	{\includegraphics[trim=10 15 10 15, clip,width=0.45\columnwidth]{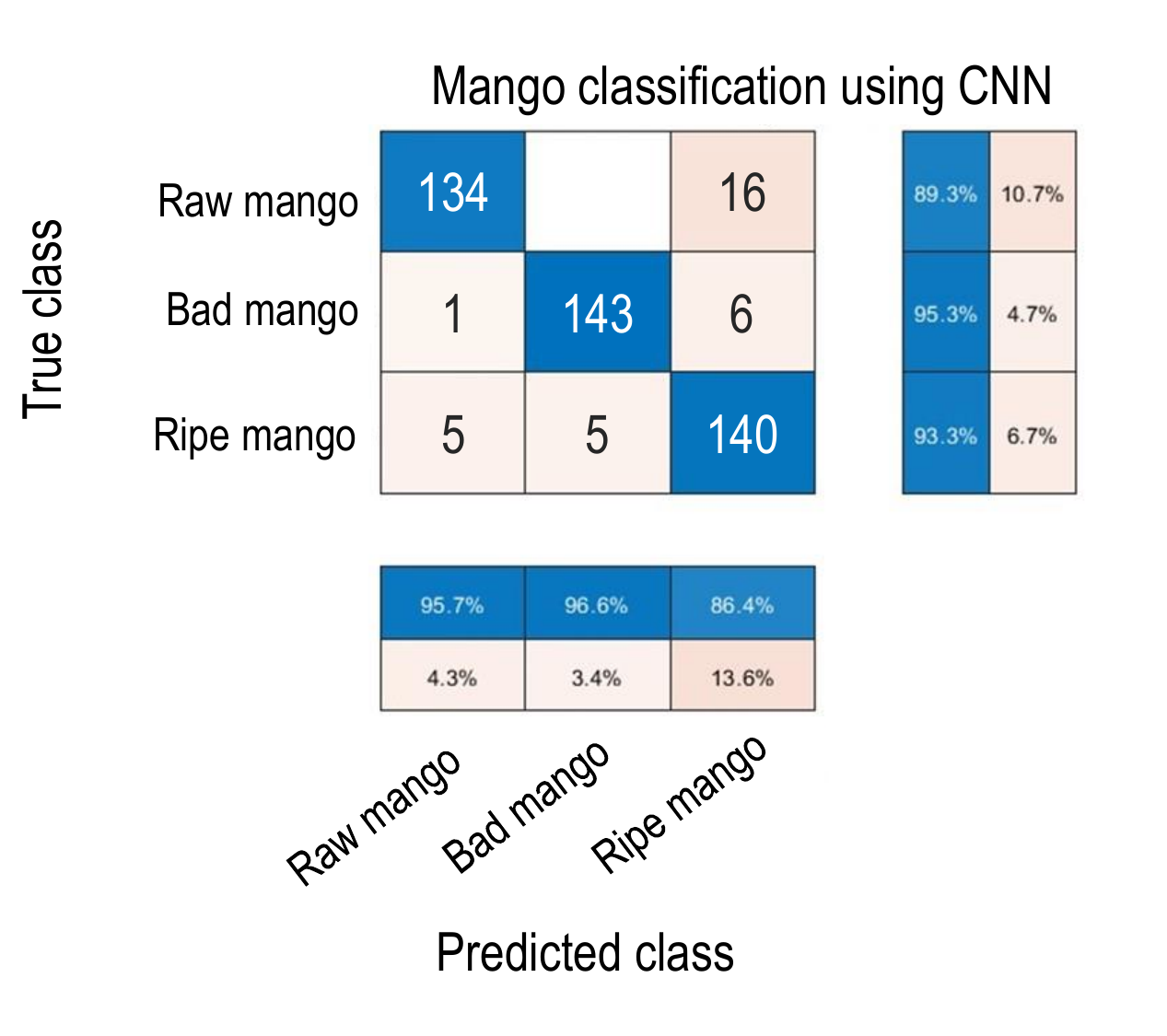}} 
	\hfill 
	\caption{Ripeness classification of mangoes using the Resnet-18 model.} 
	\label{fig23} 
\end{figure}

A model based on Resnet-18 was trained to classify mangoes into three ripeness stages: raw, ripe, and rotten.
The model achieved a validation accuracy of $\SI{89.51}{\percent}$, with no evidence of overfitting, indicating a strong generalization capability.
\Cref{fig23}(a) shows a representative result: the model assigns a probability of $\SI{91.1}{\percent}$ to the ``ripe mango'' class, with significantly lower values for the other categories.
This distribution indicates that the model has successfully learned to identify distinctive visual patterns, even in samples with ambiguous features. 
The confusion matrix\footnote{Values on the main diagonal represent correct classifications; off-diagonal values correspond to classification errors, including false positives and false negatives.} shown in \Cref{fig23}(b) supports this outcome: the ``bad mango'' class reached $\SI{95.3}{\percent}$ accuracy, the ripe mango class $\SI{93.3}{\percent}$ and the ``raw mango'' class $\SI{89.3}{\percent}$, demonstrating clear category differentiation.
This consistent performance confirms ResNet-18’s suitability as the base model for ripeness classification.

As part of the comparative analysis, alternative architectures such as ResNet-50, ResNet-101, GoogLeNet, and AlexNet were evaluated across five independent tests.
\Cref{tb8} summarizes the resulting accuracies.
It is observed that the less complex networks—Resnet-18 and AlexNet—produced better and more stable results than the more complex architectures—ResNet-50 and ResNet-101—even though validation accuracies were relatively low across the board.
This suggests that for this relatively simple task, smaller models are more effective and less prone to overfitting.

\begin{table}[htb] 
	\caption
	{\label{tb8} Model Accuracy.} 
	\begin{center} 
		\begin{tabular}{|c|c|c|c|c|c|} 
			\hline Model & Test 1 & Test 2 & Test 3 & Test 4 & Test 5\\ 
			\hline ResNet-50 & $\SI{86.67}{\percent}$ & $\SI{78.33}{\percent}$ & $\SI{43.33}{\percent}$ & $\SI{78.33}{\percent}$ & $\SI{33.33}{\percent}$\\ 
			\hline Resnet-18 & $\SI{93.33}{\percent}$ & $\SI{55.00}{\percent}$ & $\SI{33.33}{\percent}$ & $\SI{83.33}{\percent}$ & $\SI{16.67}{\percent}$\\ 
			\hline ResNet-101 & $\SI{81.67}{\percent}$ & $\SI{30.00}{\percent}$ & $\SI{40.00}{\percent}$ & $\SI{85.00}{\percent}$ & $\SI{33.33}{\percent}$\\ 
			\hline GoogLeNet & $\SI{76.67}{\percent}$ & NaN\footnotemark & NaN & $\SI{90.00}{\percent}$ & $\SI{33.33}{\percent}$\\ 
			\hline AlexNet & $\SI{73.33}{\percent}$ & NaN & $\SI{33.33}{\percent}$ & $\SI{46.67}{\percent}$ & $\SI{33.33}{\percent}$\\ 
			\hline 
		\end{tabular} 
	\end{center} 
\end{table}
\footnotetext{Not a Number. This error may arise from exploding gradients, invalid data (such as infinite values), or optimization issues where the algorithm fails to reach a global minimum.}

\begin{figure}[htb] 
	\centering 
	\subfloat[AlexNet] {\includegraphics[trim=10 15 10 15,clip,width=0.45\columnwidth]{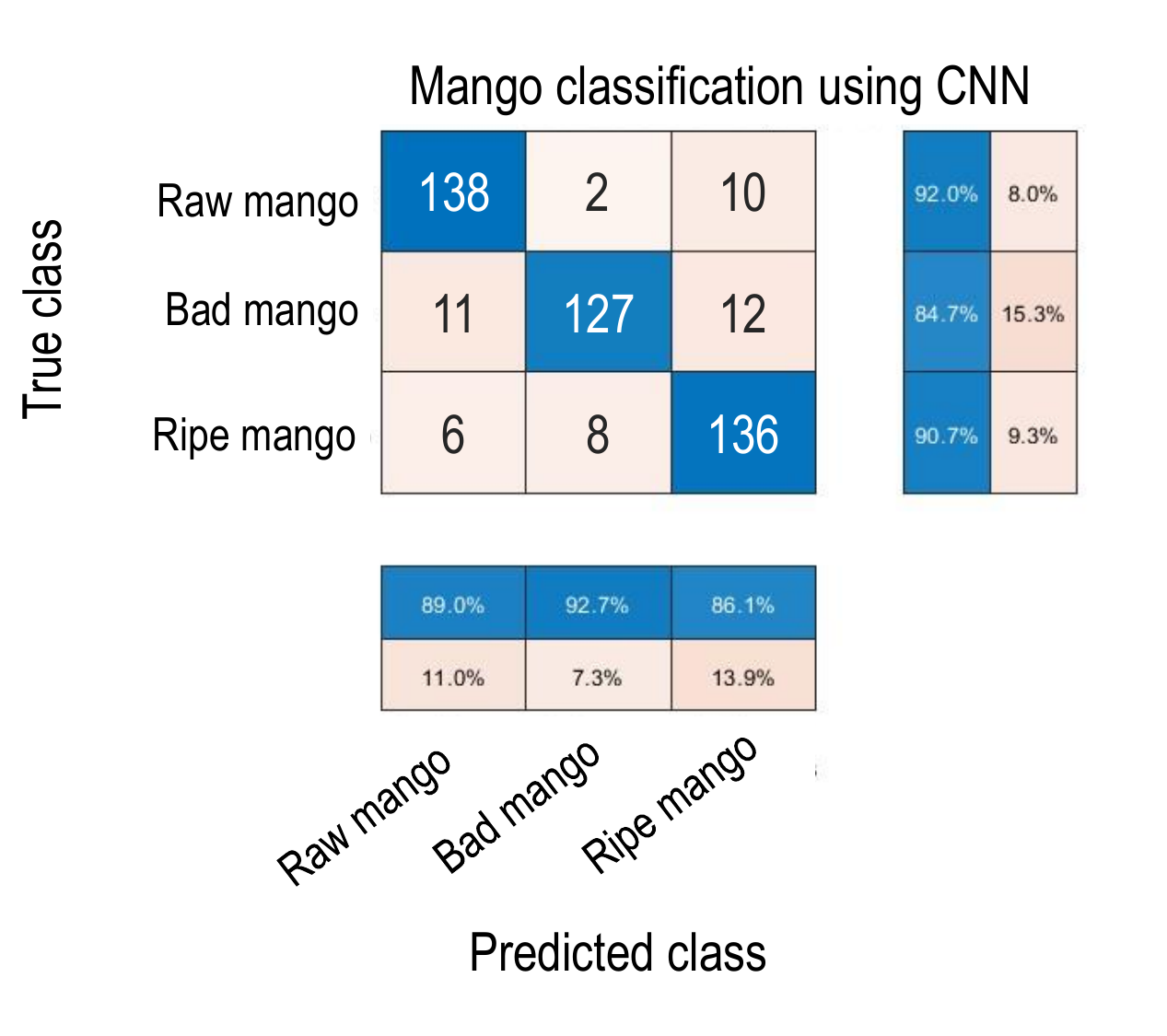}}
	\hfill 
	\subfloat[GoogLeNet] 
	{\includegraphics[trim=10 15 10 15, clip,width=0.45\columnwidth]{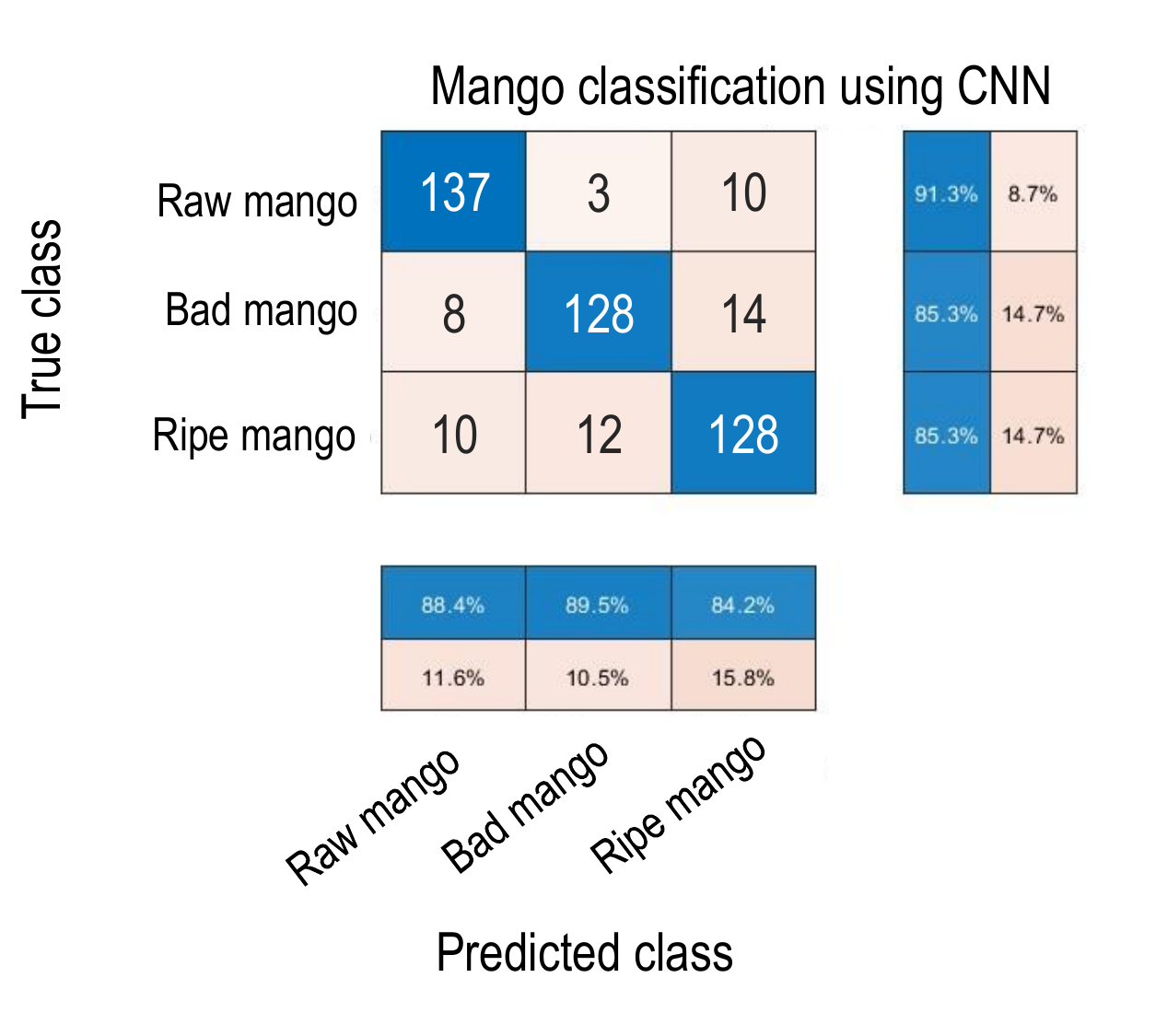}} 
	\caption{Confusion matrices for AlexNet and GoogLeNet models.} 
	\label{fig24_25_26} 
\end{figure}

AlexNet and GoogLeNet were selected alongside Resnet-18 for a performance comparison, due to a favorable balance between accuracy, stability, and generalization.
Their respective confusion matrices are presented in \Cref{fig24_25_26}.
For AlexNet (\Cref{fig24_25_26}(a)), strong performance was observed in the ``raw mango'' class ($\SI{92.0}{\percent}$), though higher confusion appeared in the ``bad mango'' category ($\SI{84.7}{\percent}$), indicating limitations when classifying samples with ambiguous visual traits.
GoogLeNet (\Cref{fig24_25_26}(b)) showed moderate performance overall, with frequent errors in the ``ripe mango'' class ($\SI{85.3}{\percent}$), likely due to subtle image variations or mild overfitting. 
Collectively, the obtained results and matrix patterns confirm that Resnet-18 provides the best balance of accuracy, robustness and computational efficiency.
As such, Resnet-18 was selected as the most suitable model for mango ripeness classification.

\subsection*{Training and validation of the disease classification model}

\begin{figure}[htb] 
	\centering 
	\subfloat[Disease classification] 
	{\includegraphics[width=0.4\columnwidth]{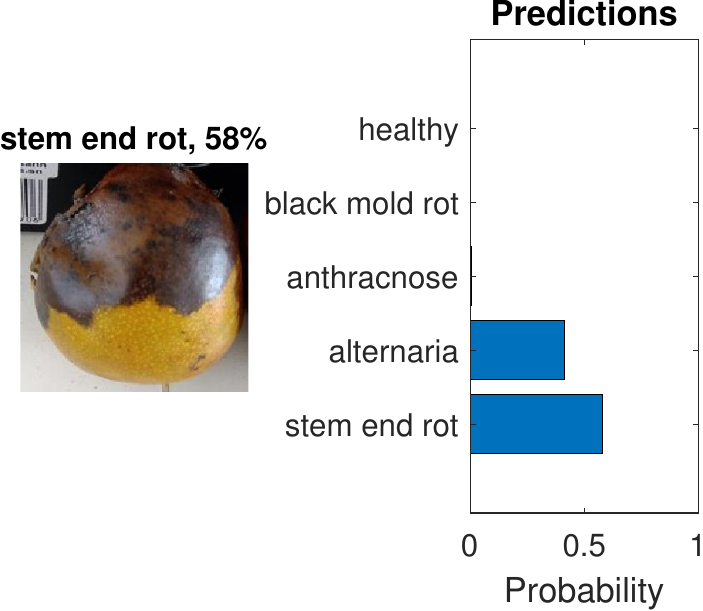}} 
	\hfill 
	\subfloat[Confusion matrix of the model] 
	{\includegraphics[width=0.45\columnwidth]{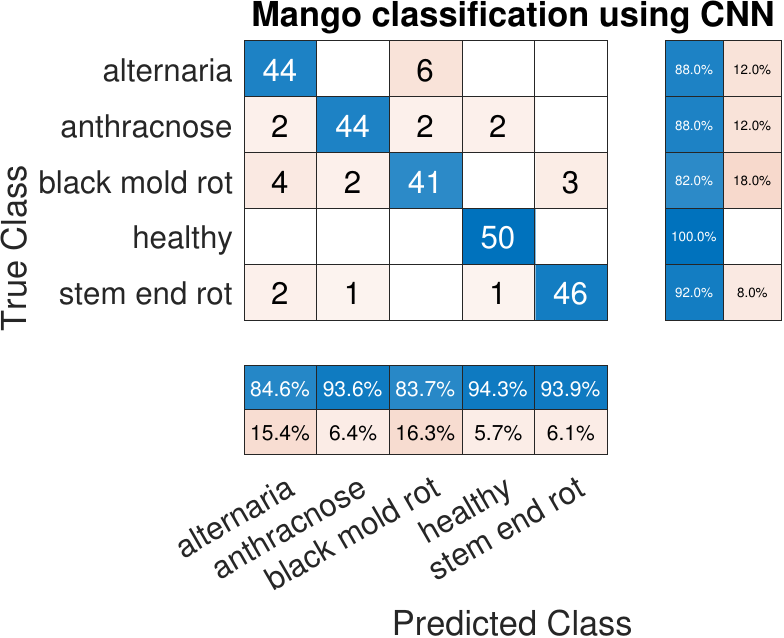}} 
	\caption{Classification of mango diseases using a dedicated disease model.} 
	\label{fig27} 
\end{figure}

A Resnet-18 model was trained to classify mangoes into five categories: alternaria, anthracnose, black mold rot, stem end rot and healthy. 
In \Cref{fig27}(a), a prediction example is shown in which the model assigns a probability of $\SI{58}{\percent}$ to the ``stem end rot'' class, consistent with the visible characteristics in the image.
The low probabilities assigned to the remaining classes indicate that the model can distinguish between pathological symptoms and normal conditions correctly.

The confusion matrix shown in \Cref{fig27}(b) demonstrates solid performance, with accuracies above $\SI{88.0}{\percent}$ for all categories and relatively low misclassification rates.
The ``healthy'' class was correctly classified in $\SI{100}{\percent}$ of cases, indicating a clear distinction between healthy and diseased fruits.
The ``alternaria'', ``anthracnose'' and ``black mold rot'' categories showed some confusion among themselves, which was expected given the morphological similarity of their symptoms.
These results confirm that the dedicated disease classification model is reliable, especially for classes with well-defined visual patterns.

\begin{figure}[htb] 
	\centering 
	\subfloat[Disease classification] 
	{\includegraphics[width=0.4\columnwidth]{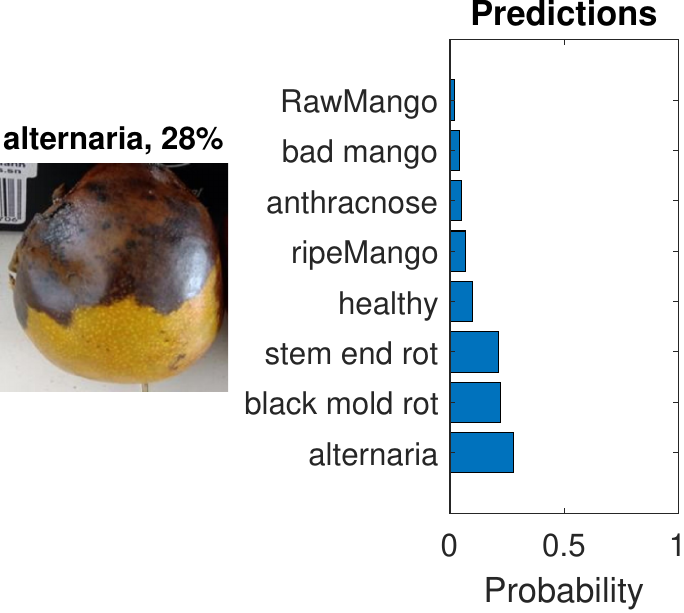}} 
	\hfill 
	\subfloat[Confusion matrix of the model] 
	{\includegraphics[trim=10 15 10 15, clip,width=0.55\columnwidth]{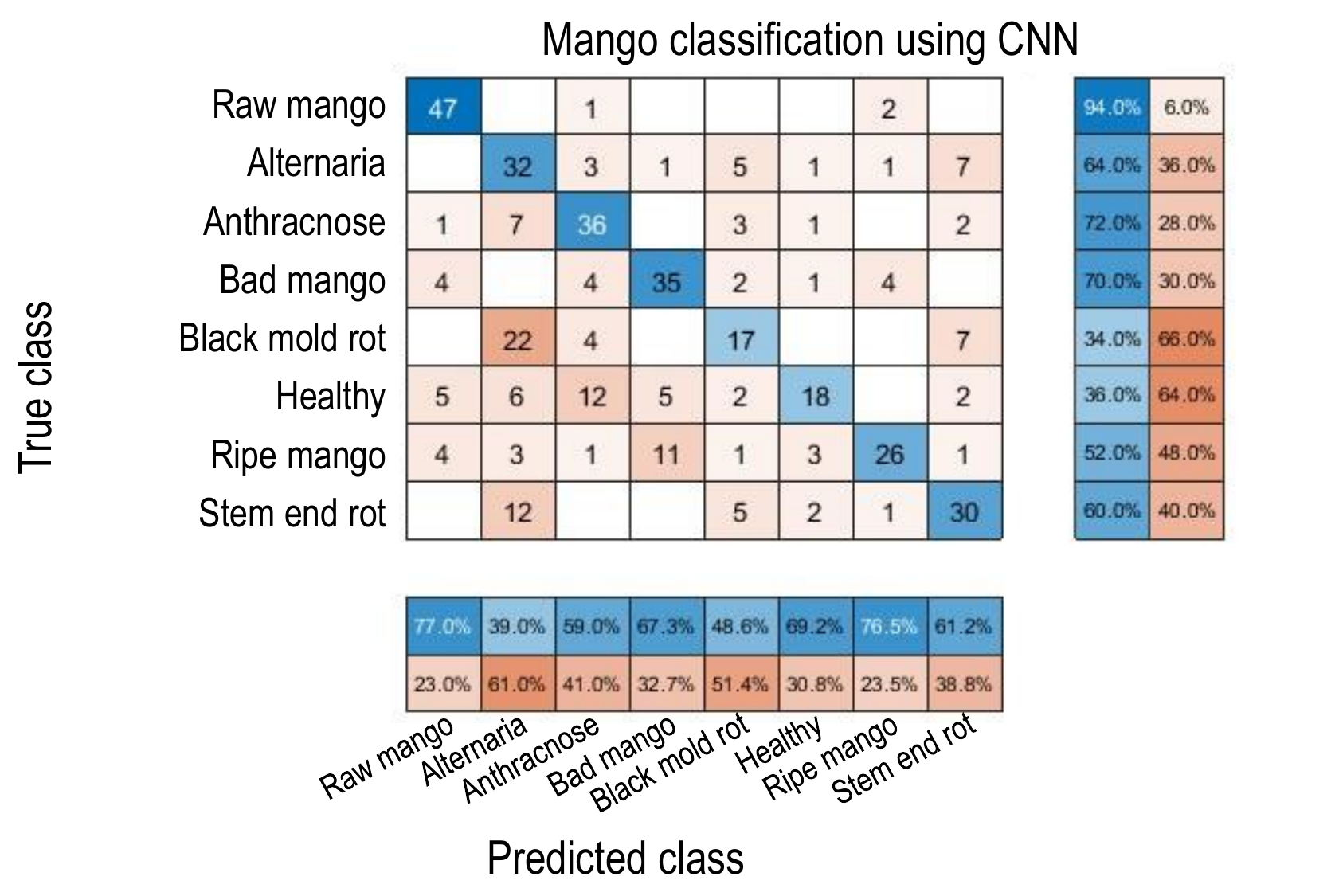}} 
	\caption{Classification of mangoes using a combined model with multiple categories.} 
	\label{fig29} 
\end{figure}

Subsequently, a combined model was trained to classify both diseases and ripeness stages.
\Cref{fig29}(a) shows a case where the model assigns the highest probability ($\SI{28}{\percent}$) to ``alternaria,'' although visually the fruit exhibits more distinctive symptoms of ``stem end rot''.
This prediction reflects higher uncertainty, likely due to overlapping features between classes.

The confusion matrix shown in \Cref{fig29}(b) confirms this trend: a greater dispersion of errors is observed, particularly among diseases with similar symptoms and between diseased and ripe fruits.
The ``alternaria'', ``anthracnose'' and ``black mold rot'' classes show frequent cross-classification errors, as do ``bad mango'' and ``stem end rot''.
Additionally, in some cases, diseased fruits were classified as ``ripe mango'', suggesting that the model may interpret certain symptoms as part of the natural ripening process.

This behavior indicates that including multiple categories in a single network may hinder discrimination between classes with visually similar symptoms, affecting the model’s specialization.
Although the combined model yields functional results, its overall performance is lower in terms of precision and interpretability.
Therefore, it is not recommended to address ripeness and disease classification within a single model simultaneously. 
Dividing the tasks allows for more targeted treatment of each problem, reducing complexity and improving the clarity of the results.

\subsection*{Visual application}

\begin{figure}[htb] 
	\centering 
	\includegraphics[trim=0 10 0 35, clip,width=0.7\columnwidth]{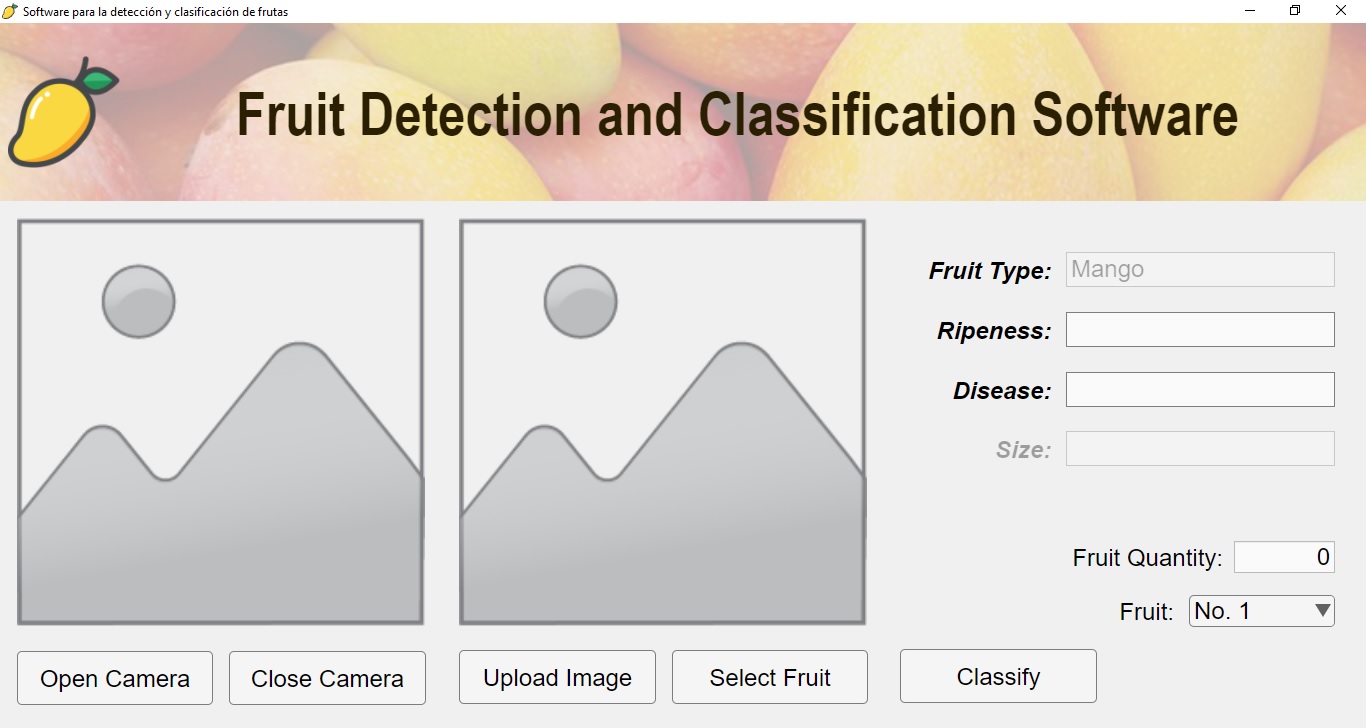} 
	\caption{Visual interface for mango classification.} 
	\label{fig15} 
\end{figure}

The developed models were integrated into a visual application, shown in \Cref{fig15}, using MatLab’s App Designer to optimize result visualization.
The application includes several functions to facilitate user interaction with images and the classification process:

\begin{itemize} 
	\item Load Image: allows the user to select images from the device. 
	\item Open/Close Camera: enables real-time image capture. 
	\item Select Fruit: applies the detection algorithm to identify mangoes within the image. 
	\item Classify: runs the classification model on the selected image fragment. 
\end{itemize}

MatLab App Designer was chosen due to its convenience for designing and programming visual interfaces.
Additionally, using the same software for both model training and the visual application helps avoid potential compatibility issues in future implementations.

\section{CONCLUSIONS}

The results introduced in this paper confirm that \gls{cnn} are effective for mango detection and classification, integrating a cascade object detector and Resnet-18 into a visual application developed with MatLab App Designer.
This system enhances classification accuracy, reduces human error, and eliminates the need for destructive methods. 
Furthermore, the developed interface facilitates access to information and supports the export of mangoes by meeting international quality standards.
Future work will consider the integration of this solution within a real-time system pipeline.
Examples include the monitoring of crops with drones, where integrated cameras support the collection of images that this \gls{cnn}-based autonomous system can evaluate.

\printbibliography[title=5.\hspace{10pt} References]

\subsection*{AUTHORS SECTION}

Beatriz Díaz Peón: Professor at the Havana University of Technology ``José Antonio Echeverría'' (CUJAE), Havana, Cuba. Automation Engineer. Member of the Image Processing Research Group. Teaching category: Instructor-in-training.

Jorge Torres Gómez: Senior Researcher at the School of Electrical Engineering and Computer Science, TU Berlin, Berlin, Germany. His research interests are in the fields of 6G wireless technologies, molecular communications, and teaching methodologies. ORCID ID 0000-0001-9523-048X 

Ariel Fajardo Márquez: Professor at the Havana University of Technology ``José Antonio Echeverría'' (CUJAE), Havana, Cuba. Master of Science Automation Engineer. Member of the Image Processing Research Group. Teaching category: Assistant Professor.

\subsection*{CONFLICT OF INTEREST}

The authors declare that there are no conflicts of interest regarding the content of this article.

\subsection*{CONTRIBUTIONS OF THE AUTHORS}

\begin{itemize}
	\item Author 1: Beatriz Díaz Peón - $\SI{60}{\percent}$ Conceptualization, writing and development of the article and critical review of each draft version.
	\item Author 2: Jorge Torres Gómez - $\SI{20}{\percent}$ Conceptualization, critical review of each draft version and approval of the final version to be published.
	\item Author 3: Ariel Fajardo Márquez - $\SI{20}{\percent}$ Conceptualization, critical review of each draft version and approval of the final version to be published.
\end{itemize}

\begin{figure}[htb]
\includegraphics{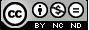}
\end{figure}


\end{document}